\documentclass[11pt]{article}

\usepackage[preprint]{acl}

\usepackage{times}
\usepackage{latexsym}

\usepackage[T1]{fontenc}

\usepackage[utf8]{inputenc}
\usepackage{microtype}
\usepackage{inconsolata}
\usepackage{graphicx}
\usepackage{amsmath}
\usepackage{booktabs}
\usepackage{float}
\usepackage{makecell}

\title{Formal Adequacy Index\thanks{An article based on Chapter 6 of \citeposs{Patreucha} Master’s thesis.} \\ (based on Polish-language material)}

\author{Paweł Patreucha \\
  Univeristy of Wrocław \\}

\begin{document}
\maketitle
\begin{abstract}
Linguistic formality is one of the dimensions of stylistic variation in speech; however, simply measuring its level does not allow us to determine whether the chosen mode of communication is appropriate in a given situation. The aim of this article is to present the concept of a formal adequacy index (based on Polish-language material), which enables a quantitative assessment of how well the formality of a text matches the expectations arising from the communicative context. The proposed approach is based on the separate modelling of observed formality, determined on the basis of the linguistic features of the text, and expected formality, determined on the basis of the characteristics of the communicative situation. Adequacy is assessed by comparing these two measures, taking into account contextual tolerance thresholds, which may differ for a lack of formality and an excess of formality. The potential applications of the index are illustrated by an analysis of Polish-language emails containing apologies for being late, written by human language users or generated by a large language model. The implementation presented here is preliminary in nature and requires further calibration and empirical validation. The main contribution of this article is the proposal of a general methodological framework for the quantitative study of formal adequacy in various communicative contexts.
\end{abstract}

\section{Introduction}

The way in which an text is formulated is linked to the communicative situation in which it is produced. The choice of linguistic means may depend, amongst other things, on the purpose of communication, the genre of the text, the relationship between the speaker and the listener, and the norms prevailing in a given environment. One aspect of this variation is linguistic formality. Although its level can be described on the basis of the text’s characteristics, assessing the appropriateness of the style used requires taking into account the expectations arising from the communicative context. Consequently, a text with a specific level of formality may be appropriate in one situation, whilst in another it may be perceived as too formal or too informal.

\par The distinction between the formality of a text and the adequacy of its level to a given communicative situation provides the starting point for proposing a distinct measurement approach. In this article, \textbf{formality adequacy} is understood as the degree to which the observed level of formality of a text matches the level expected in a given communicative context. This approach assumes that an assessment of appropriateness requires consideration of two elements: the linguistic properties of the text under analysis and the characteristics of the communicative situation, which shape expectations regarding how that text should be formulated. 

\par The index presented here serves as a methodological framework that can be operationalised in various ways, depending on the type of texts analysed, the information available about the participants in the communication, and the research assumptions adopted. The selection of linguistic features, context-describing variables, the weights assigned to them, and tolerance thresholds requires justification and verification on a case-by-case basis. In particular, expert-based solutions can serve as a starting point for constructing a model, which can then be developed and calibrated on the basis of empirical data.

\par To demonstrate the potential applications of the proposed framework, we also present an illustrative case study analysing emails containing apologies for lateness, addressed to recipients with diverse social and relational characteristics. This study employs an expert-based operationalisation of the expected level of formality alongside a set of measurable linguistic features to determine observed formality. Rather than serving as a universal solution, this example is intended to illustrate how the index operates in practice.

\section{Related works}
One of the fundamental principles of pragmatics and sociolinguistics is the belief that the way language is used cannot be considered in isolation from the communicative situation (\citet{Hymes1972}, \citet[p 31-35, 108-126]{halliday1978language}, \citet[p 6-11]{biber2009register}, \citet{pescuma2023situating}). Research indicates that the assessment of a statement should take its communicative context into account. The same form of expression may meet expectations in one situation, whilst in another it may be perceived as inappropriate. This applies to various aspects of linguistic style, including its degree of formality.

Formality is one aspect of linguistic variation; however, no single, widely accepted method for measuring the formality of texts has yet been developed (\citet{sigley1997text}, \citet{peterson2011email}, \citet{heylighen1999formality}, \citet{li2013comparing}, \citet{pavlick2016empirical}). In the Polish linguistic tradition, variation across texts is addressed, for instance, within the framework of research on linguistic varieties and functional stylistics (\citet[p 14-16, 96-105, 162-169]{zdunkiewicz2008wyklady}). Formality refers to specific linguistic features of a text, whereas appropriateness requires these features to align with the expectations shaped by the context.

Text formality can also be examined from the perspective of Genre Studies. Research into communicative genres emphasises that recurring communicative goals and established conventions shape both the structure of a text and the choice of linguistic features (\citet[p 58]{swales1990genre}, \citet[p 3-12]{bhatia1993analysing}, \citet[p 15-19]{biber2009register}). In the Polish linguistic tradition \citet{wojtak2014gatunkach} points out that the genre pattern encompasses, amongst other things, pragmatic and stylistic dimensions, and that its specific manifestations may exhibit considerable variation. This does not, however, mean that each genre corresponds to a single, fixed level of formality. Rather, genre conventions may help to shape expectations regarding the way in which texts are formulated. An empirical example of how genre influences the way a statement is formulated is provided by \citeposs{zasko2014apologizing} research into the act of apologising in Polish farewell letters.
From the perspective of the proposed index, the type of text may therefore be one of the factors influencing the expected level of formality, alongside social relationships and other aspects of the communicative context.

One of the key factors shaping stylistic choices is the nature of the audience. In formulating his audience design theory, \citet{bell1984language} points out that stylistic variation can result from adapting language use to the audience. This phenomenon of adapting to one's interlocutors is also central to Communication Accommodation Theory (\citet{giles1991accommodation}). In the context of the Polish language, the relationship between politeness norms, the nature of interpersonal relationships and the communicative situation is discussed, amongst others, by \citet[p 11–34, 70–76]{marcjanik2007grzecznosc} and \citet{pachowicz2014bym}. It should be noted, that the influence of individual recipient characteristics cannot be viewed as straightforward or independent of other factors. Expectations regarding text style also depend on the broader context, the type of communication, established conventions, and the existing relationship between participants. These considerations justify treating the expected level of formality as a context-dependent variable. At the same time, they suggest that determining this level must account for the co-occurrence and interplay of various social and situational factors. Consequently, establishing the expected level of formality requires adopting a specific contextual model, whose detailed structure may vary depending on the type of communication under study.


\section{Formal Adequacy Index}
\subsection{General Assumptions of the Index}
Existing research indicates that text formality can be analysed both as an inherent linguistic property of a text and as a phenomenon dependent on the communicative situation. This distinction forms the basis of the proposed Formal Adequacy Index, which aims to quantitatively evaluate the degree to which a text's formality aligns with the expectations shaped by its context of use.

\par The construction of the index is based on the assumption that observed formality and expected formality can be modelled independently using distinct sets of information. The former describes the level of formality observed in a text based on selected linguistic features. The latter refers to the level of formality expected in a given communicative situation, determined by its characteristics, particularly the relationship between participants and prevailing linguistic norms.

\par In general terms, both quantities can be expressed as follows:
\begin{equation}
   E_i=g(C_i),
   \label{eq:expected formality}
\end{equation}
\begin{equation}
    F_i = f(T_i),
    \label{eq:real formality}
\end{equation}
where $E_i$ denotes the expected level of formality of the $i$-th text, $F_i$ denotes its observed level of formality, $C_i$ describes the communicative context, and $T_i$ is the set of linguistic features of the text. The functions $f$ and $g$ represent the adopted methods for estimating both quantities. Both the observed formality model and the communicative expectations model can take various forms, depending on the type of texts analysed, the language, and the available data. In particular, it is possible to employ models based on expert rules, statistical methods, or other empirical procedures.

\par The assessment of adequacy is based on a comparison of both formality levels. Assuming they are expressed on a comparable scale, we can define their difference as:
\begin{equation}
  D_i = F_i - E_i.
  \label{eq:difference}
\end{equation}
Positive values of $D_i$ indicate an excess of formality relative to the expected level, whereas negative values denote a deficit. A value close to zero corresponds to a situation in which observed formality closely aligns with expected formality.

\par However, the mere presence of a difference does not necessarily indicate text inadequacy. In certain communicative situations, deviations from the expected level of formality may be permissible. Moreover, the acceptable range of formality deficit and excess does not have to be symmetric. For this reason, the proposed index allows for the introduction of tolerance thresholds dependent on the communicative context.

\subsection{Model of the Expected Level of Formality}
The expected level of text formality stems from the conditions of the communicative situation in which a text is produced. Determining this level requires accounting for factors that shape linguistic norms and expectations regarding language use. These factors may include, among others, the social relationships between participants, the communicative goal of the text, and the conventions characteristic of a given domain or communicative genre.

\par For the purposes of the proposed index, the expected level of formality is treated as a quantity derived from the adopted communicative context. This approach does not assume the existence of a single, objective level of formality appropriate for every situation. Instead, the value of expected formality is the result of a specific operationalisation of communicative norms, which depends on the factors considered and the way their influence is specified.

\par In accordance with the adopted assumptions, the model of the expected level of formality can be expressed as in Equation \eqref{eq:expected formality}. The functional form of $g$ is not fixed a priori and may depend on the available data and the nature of the communication analysed. One possible approach is an additive model, in which the expected level of formality is computed as a weighted sum of contextual factors:
\begin{equation}
  E_i=\sum_{k=1}^{K}\beta_k M_{ik},
  \label{eq:expected_sum}
\end{equation}
where $M_{ik}$ denotes the value of the $k$-th contextual variable for the $i$-th text, $\beta_k$ determines the direction and relative strength of its influence on the expected level of formality, and $K$ is the number of variables included.

\par The additive model provides a relatively simple method for aggregating information about the communicative context. However, it assumes that the contribution of individual factors can be considered independently. In more complex communicative situations, it may be warranted to account for interactions between variables or to employ non-linear models. 

\par The weights assigned to individual factors can be determined by experts, derived from empirical studies, or calibrated using assessments from language users. The choice of weighting procedure should depend on the research objective and available data. Regardless of the adopted procedure, it is necessary to justify the selection of variables and evaluate the extent to which the constructed model reflects communicative expectations in the analysed domain.

\subsection{Model of the Observed Level of Formality}
The observed level of formality refers to the observable linguistic properties of a text. Unlike expected formality, which is determined on the basis of the communicative context, estimating observed formality relies on analysing the features of the text itself. Formality can manifest across various levels of text organisation, including lexical choice, syntactic constructions, as well as forms of address and politeness markers.

\par As with expected formality, the proposed framework does not assume a single, universal measurement method. The selection of linguistic features should depend on the language analysed, the type of text, and the adopted definition of formality. It is also important to recognise that individual textual properties do not necessarily indicate a specific level of formality unambiguously, and their significance may vary depending on the context and communicative genre.

\par In general terms, the model of the observed level of formality can be expressed as in Equation \eqref{eq:real formality}. One possible way of determining observed formality is to construct an index based on selected linguistic indicators. In this case, the level of formality can be defined as a weighted sum:
\begin{equation}
  F_i=\sum_{j=1}^{J}\alpha_jZ(X_{ij}),
  \label{eq:real_sum}
\end{equation}
where $X_{ij}$ denotes the value of the $j$-th linguistic feature for the $i$-th text, $Z$ is a standardising function, $\alpha_j$ represents the weight assigned to a given feature, and $J$ denotes the number of indicators included.

\par Standardisation enables the aggregation of features measured on different scales; however, it does not determine their suitability for measuring formality. The inclusion of individual indicators requires separate theoretical or empirical justification.

\par An alternative to indices based on predefined features is offered by statistical and machine learning models, in which the relationship between textual properties and perceived formality is derived from empirical data. Human ratings can also be utilised either as a direct source of information regarding a text's level of formality or as a benchmark for validating automated models.

\par Regardless of the method employed, the measurement of observed formality must allow for direct comparison with the expected level of formality. This requires ensuring the comparability of scales for both quantities, which may necessitate additional calibration. Only when this condition is met can the difference between observed and expected formality be interpreted as a valid basis for assessing formal adequacy.

\subsection{Formal Adequacy Criterion}
The assessment of formal adequacy is based on comparing the observed level of formality of a text with the level expected in a given communicative context. Assuming that both quantities are expressed on a comparable scale, the difference between them can be defined as in Equation \eqref{eq:difference}. The sign of the difference $D_i$ determines the direction of the mismatch. Positive values indicate an excess of formality relative to the expected level, whereas negative values denote a deficit. A value of zero corresponds to an exact match between the two levels.

\par However, the magnitude of the difference alone does not determine whether a text should be considered adequate, as minor deviations from the expected level of formality may be permissible in certain communicative situations. Therefore, we introduce tolerance thresholds that define the range of deviations deemed acceptable. This range may depend on the communicative context, and the permissible deficit and excess of formality need not be symmetric. The adequacy criterion can thus be formulated as:
\begin{equation}
  -\delta_{c_i}^{-}\leq D_i\leq\delta_{c_i}^{+},
  \label{adeq_criterion}
\end{equation}
where $c_i$ denotes the communicative context of the $i$-th text, $\delta_{c_i}^{-}$ is the permissible formality deficit, and $\delta_{c_i}^{+}$ corresponds to the permissible formality excess. Both parameters take non-negative values.

\par Introducing separate thresholds stems from the assumption that the consequences of insufficient and excessive formality may vary depending on the situation. For example, in communication requiring a high level of formality, adopting an overly casual style may constitute a more serious deviation from expectations than using a slightly more formal one. Conversely, in informal settings, the opposite pattern may occur, where excessive formality is less acceptable than a slight formality deficit. These relationships are hypothetical in nature and require empirical verification within specific communicative contexts.

\par Tolerance threshold values can be set by experts or derived from empirical evaluations. It is also possible to apply uniform thresholds across the entire corpus of texts; however, accounting for their context dependence allows for more flexible modelling of communicative expectations. The choice of procedure should be guided by the research objective, the nature of the material analysed, and the available data.

\par Based on the adopted criterion, a binary formal adequacy variable can be defined as:
\begin{equation}
  A_i=
\begin{cases}
1, & \text{if }-\delta_{c_i}^{-}\leq D_i\leq\delta_{c_i}^{+},\\
0, & \text{otherwise}.
\end{cases}
\label{eq:binary_adeq}
\end{equation}
A value of $A_i=1$ indicates that the level of formality falls within the adopted tolerance range, whereas $A_i=0$ signifies that one of its bounds has been exceeded. Defined in this way, the variable enables, among other things, comparing the proportions of texts meeting the adequacy criterion across different groups and modelling the probability of satisfying it.

\par Binary classification is not, however, the only possible way of presenting the outcome. Depending on the analytical objective, one may retain the raw difference $D_i$, which captures both the direction and magnitude of the deviation, or construct a continuous measure that accounts for the extent to which the permissible thresholds are exceeded. Such approaches make it possible to distinguish texts that deviate only slightly from expectations from those exhibiting a substantial mismatch.

\par The proposed criterion thus provides a general framework for the quantitative assessment of formal adequacy. Its specific implementation, including the choice of output representation and tolerance threshold values, depends on the adopted operationalisation and should be subject to theoretical justification and empirical verification.

\subsection{Operationalisation and Potential Extensions of the Index}
The presented concept of the Formal Adequacy Index provides a general methodological framework whose application requires adaptation to the nature of the material analysed and the research objective. In particular, it is necessary to specify the methods for measuring observed formality, modelling expected formality, and determining the permissible range of deviations between the two quantities. Individual components can be operationalised in various ways while maintaining the core assumption that the assessment of adequacy relies on comparing the linguistic properties of a text with the expectations arising from its communicative context.

\par The first stage of operationalisation is to define the scope of the communicative situations analysed and select the variables used to describe them. The selection of these variables should account for the specifics of the language under study, the text genre, and the relationships between participants. Analogously, it is necessary to identify linguistic features that can serve to determine the observed level of formality within the chosen domain of application. Both the selection of contextual variables and linguistic indicators should be grounded in theoretical premises or findings from previous research.

\par A subsequent issue involves specifying the model parameters. The weights assigned to variables and the values of tolerance thresholds can be set by experts or derived from empirical data. While an expert-driven approach enables the construction of an initial version of the index even when data availability is limited, parameters established in this manner remain assumptions requiring further verification. An alternative is to utilise ratings from language users, which can serve to calibrate the models and define the ranges of formality deemed acceptable in specific contexts.

\par A crucial condition for applying the index is ensuring the comparability of the scales for observed and expected formality. Because both quantities can be derived from distinct sets of information using different methods, a direct comparison between them is not always warranted. Depending on the adopted operationalisation, it may be necessary to apply appropriate transformations or calibration procedures to allow the difference between formality levels to be meaningfully interpreted. Merely rescaling the variables to a similar numerical range does not, however, guarantee that the resulting scales are interpretatively equivalent.

\par An important component of evaluating the proposed approach should be a sensitivity analysis of the results with respect to the underlying assumptions. Specifically, it is possible to examine the extent to which changes in variable weights, aggregation methods, and tolerance threshold values affect the final adequacy assessment. Such an analysis makes it possible to determine the robustness of the obtained results and identify the model components to which they are particularly sensitive.

\par Empirical validation of the index remains a separate issue. One possible approach is to compare its outputs with independent assessments of formal adequacy provided by language users. These evaluations should take into account both the content of the text and information about the communicative situation in which the text is intended to function. This would make it possible to verify whether the index reflects how recipients perceive formal appropriateness, as well as to identify contexts in which the adopted operationalisation requires modification.

\par Potential avenues for future development also include expanding the set of linguistic features and contextual variables considered, applying more complex models of the relationships between them, and developing alternative ways of presenting the outputs. Particularly promising is the possibility of deriving context-dependent tolerance thresholds from empirical data rather than setting their values a priori. This would allow for a gradual reduction in the index's reliance on subjective assumptions made during its construction.

\section{Illustrative Implementation}
To demonstrate the potential applications of the proposed index, we present an example of its use in the analysis of Polish-language emails. The implementation employed relies on an expert-based model of the expected level of formality and a set of measurable linguistic features used to determine observed formality. The presented example constitutes one possible realisation of the framework described in the previous section.

\subsection{Characteristics of the Research Material}
The analysis was conducted on a corpus of 420 emails containing apologies for lateness, addressed to recipients with diverse social and relational characteristics. The study included both emails written by human language users and those generated by a large language model (LLM).

\par The research material from human language users was collected via an anonymous online form. Participation in the study was voluntary, and participants were informed that the submitted emails would be used for scientific purposes. The form collected only a limited scope of participant information, including self-reported first names or pseudonyms, age, and general study-related metadata. This data collection procedure was designed to minimise the possibility of identifying participants. The analysis focused on the linguistic properties of the collected texts rather than the individual linguistic competence of their authors.

\par The assembled corpus enabled a comparison of how text formality is adapted to the communicative context, with particular emphasis on differences between emails from both sources. The context of each email was described using a set of metadata characterising the recipient and their relationship with the sender.

\par The aim of the presented application was to examine the feasibility of using the index to quantitatively describe the alignment of an email's formality with the expectations arising from the specified characteristics of the communicative context.

\subsection{Operationalisation of Expected Formality}
The expected level of formality was determined based on five metadata categories: the recipient's role relative to the sender, the recipient's age relative to the sender, the degree of involvement in the interpersonal relationship, the recipient's attitude towards politeness norms, and their level of education.

\par Individual metadata values were assigned expert weights from the set $\{-2,-1,0,+1,+2\}$, determining the direction and relative strength of their influence on the expected level of formality. The greatest weight was assigned to the recipient's role, which was treated as the primary factor reflecting the nature of the social relationship and communicative distance.

\par For each email, the expected level of formality was computed according to the additive model given in Equation \eqref{eq:expected_sum}, where $M_{ik}$ denotes the value of the $k$-th contextual variable for the $i$-th email, and $\beta_k$ corresponds to its assigned weight. The resulting index was ordinal in nature. For further analysis, it was transformed into three categories corresponding to low, medium, and high expected levels of formality. The category boundaries were determined based on the tertiles of the empirical distribution of the index.

\par This procedure made it possible to order the analysed contexts according to the expected level of formality. At the same time, the weight values and category boundaries were either expert-defined or sample-dependent; therefore, they should not be treated as universal communicative norms.

\subsection{Operationalisation of Observed Formality and the Adequacy Criterion}
The observed level of formality was determined based on three linguistic features of the email: a text readability index, the number of politeness markers, and text length measured by word count.

\par To mitigate the influence of extreme observations, appropriate logarithmic transformations were applied to selected variables. Subsequently, all three components were standardised, and the observed formality index was defined as their sum:
\begin{equation}
  F_i=Z(R_i)+Z(P_i)+Z(L_i),
  \label{eq:rreal_formality_mail}
\end{equation}
where $R_i$ denotes the transformed readability score, $P_i$ is the number of politeness markers, and $L_i$ represents the transformed email length.

\par Constructed in this manner, the index served as a relative measure of formality within the analysed corpus. The selected indicators were treated as approximate proxies for linguistic formality rather than an exhaustive description thereof.

\par Formal adequacy was determined based on the difference between observed and expected formality, in accordance with Equation \eqref{eq:difference}. Subsequently, expert-defined tolerance thresholds dependent on the expected level of formality were adopted (see Table \ref{tab:values_tolerance}). Asymmetric deviation ranges were applied, assuming a greater tolerance for formality deficit in contexts with low formality requirements and a greater tolerance for formality excess in contexts with high formality requirements.

\begin{table}[H]
    \centering
    \begin{tabular}{lcc}
    \toprule 
    \makecell{\textbf{Expected}\\ \textbf{formality level}} & \textbf{$\delta^-$} (deficit) & \textbf{$\delta^+$} (excess) \\ 
    \midrule 
    low & 2 & 1 \\ 
    medium & 1 & 1 \\ 
    high & 1 & 2 \\ 
    \bottomrule 
    \end{tabular}
    \caption{Tolerance threshold values}
    \label{tab:values_tolerance}
\end{table}

\par An email was considered formally adequate if the difference between its observed and expected levels of formality fell within the adopted tolerance range. On this basis, a binary adequacy variable was defined, taking a value of 1 for emails meeting the criterion and 0 otherwise.

\par The adopted procedure was illustrative in nature. Specifically, the direct comparison of an index based on expert weights with a sum of standardised linguistic features was not preceded by an independent calibration of both scales. The resulting classifications should therefore be interpreted as the outputs of a specific operationalisation that requires further empirical verification.

\subsection{Results of the Illustrative Application}
In the analysed corpus, 108 out of $420$ emails met the adopted formal adequacy criterion, corresponding to approximately $26\%$ of the observations. The remaining $312$ emails were classified as inadequate under the applied operationalisation.

\par For further analysis, a logistic regression model was employed, in which the probability of meeting the adequacy criterion was modelled as a function of email authorship, the expected level of formality, and the interaction between these variables.


\par The obtained results suggest that the differences between LLM-generated emails and those written by human language users depend on the expected level of formality. In contexts with low expected formality, LLM-generated emails met the adopted criterion less frequently than human-written emails. A statistically significant shift in the effect of email source was also observed between contexts with low and medium expected formality. For the high expected formality level, no statistically significant interaction effect was found relative to the low level.
\begin{figure}[H]
  \centering
  \includegraphics[width=0.9\linewidth]{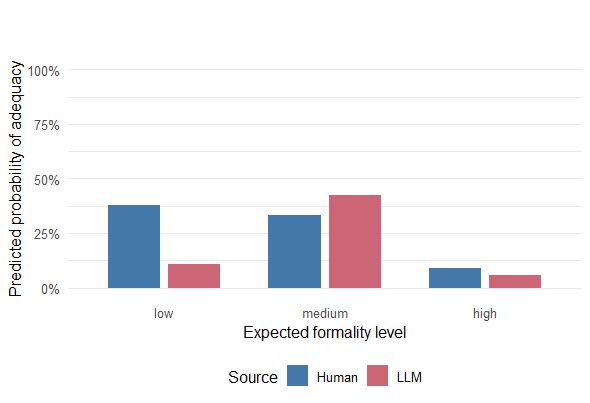}
  \caption{Predicted probability of formal adequacy.}
  \label{predicted_prob_adeq_plot}
\end{figure}

\par The presented example demonstrates the potential of using the index not only to classify individual texts, but also to conduct comparative analyses that account for the communicative context. However, the obtained results are specific to the adopted operationalisation and do not constitute independent confirmation of the index's validity. A detailed evaluation of its parameters and the alignment of its outputs with ratings from language users remains a subject for future research.

\section{Discussion}
The proposed concept of the Formal Adequacy Index offers a quantitative approach to capturing the relationship between the stylistic properties of a text and the expectations shaped by its communicative context. Unlike methods focused solely on measuring the absolute level of formality, the proposed framework enables the evaluation of how well formality aligns with a given situation. A key element of this concept is separating the modelling of observed and expected formality, as well as accounting for potential asymmetric tolerance towards their divergence.

\par Potential applications of the index include, among others, the analysis of electronic correspondence, institutional communication, and texts generated by language models. The presented illustrative implementation demonstrates the index's utility for classifying emails and comparing their adequacy across diverse communicative contexts. Due to the exploratory nature of the adopted operationalisation, however, the obtained results do not confirm the universality of the proposed tool.

\subsection{Limitations and Future Work}
The primary limitation of the proposed approach is the sensitivity of the results to how observed and expected formality are operationalised. The selection of linguistic features, contextual variables, and their assigned weights affects the values of both indicators, and consequently, the final adequacy assessment. Of particular importance in this regard is ensuring the comparability of scales, without which the interpretation of the difference between formality levels remains limited.

\par A further issue concerns determining permissible tolerance thresholds. In the presented example, these thresholds were set by experts, which allowed the functioning of the index to be demonstrated, but does not guarantee that the adopted values reflect the actual expectations of language users. Future research should therefore involve empirical parameter calibration, sensitivity analysis of the results, and comparison with independent assessments of formal adequacy.

\par Another important direction for future development is examining the applicability of the index to other text genres, languages, and communicative situations. This would make it possible to determine the extent to which the general framework can be applied independently of the specifics of a particular research corpus.

\subsection{Ethical Considerations}
\par The use of the Formal Adequacy Index requires accounting for the social and cultural nature of communicative norms. The adopted model of expectations may reflect the norms of a specific community or domain, leading to the unjustified classification of alternative modes of expression as inadequate. Therefore, the index outputs should not be treated as an objective assessment of the sender's linguistic competence, but rather as a measure of alignment with an adopted model of context and formality. Particular caution would be required if such a tool were used for the automated evaluation of specific individuals' communication.

\par A separate ethical issue concerns data collection. The presented implementation of the index was based on texts elicited for the purposes of the study rather than authentic correspondence. Validating the proposed approach on naturally occurring communicative data would be an important direction for future research; however, it would necessitate ensuring adequate privacy protection for the participants. In particular, issues such as informed consent for using correspondence for scientific purposes, personal data protection, and data anonymisation would need to be addressed. An additional challenge stems from the fact that private correspondence may contain information concerning not only the sender, but also the recipient and third parties. Consequently, the procedures for collecting and processing authentic texts should be accounted for as early as the research design stage.

\section{Conclusions}
This paper presents the concept of the Formal Adequacy Index, whose core premise is distinguishing the level of text formality from the degree to which it aligns with the communicative context. The proposed approach relies on independently estimating observed and expected formality and subsequently comparing them while accounting for a permissible range of deviations.

\par The illustrative application of the index to the analysis of Polish-language emails demonstrates the feasibility of translating this concept into a quantitative text analysis procedure. However, the presented implementation is preliminary in nature and requires further validation. The main contribution of this paper is therefore a methodological framework enabling the quantitative study of formal adequacy, whose specific formulation can be adapted to various languages, genres, and communicative contexts.


\bibliography{custom.bib}


\end{document}